\documentclass[12pt]{article}
\usepackage{graphicx}
\usepackage{authblk}
\usepackage{natbib}  
\usepackage{amsmath}
\usepackage{booktabs}
\usepackage{float}
\usepackage{multirow}
\usepackage{enumitem}
\usepackage{hyperref}

\title{An Open Benchmark Dataset for GeoAI Foundation Models for Oil Palm Mapping in Indonesia}
\author[1,2*$\dag$]{M. Warizmi Wafiq}
\author[1*$\dag$]{Peter Cutter}
\author[1,2*$\dag$]{Ate Poortinga}
\author[2]{Daniel Marc G. dela Torre}
\author[2]{Karis Tenneson}
\author[2]{Vanna Teck}
\author[2]{Enikoe Bihari}
\author[2]{Chanarun Saisaward}
\author[2]{Weraphong Suaruang}
\author[2]{Andrea McMahon}
\author[3]{Andi Vika Faradiba Muin}
\author[3]{Karno B. Batiran}
\author[3]{Chairil A}
\author[4]{Nurul Qomar}
\author[4]{Arya Arismaya Metananda}
\author[1]{David Ganz}
\author[2,5]{David Saah}
\affil[1]{RECOFTC – The Center for People and Forests, P.O. Box 1111, Kasetsart Post Office, Bangkok 10903, Thailand}
\affil[2]{Spatial Informatics Group, LLC, 2529 Yolanda Ct., Pleasanton, CA 94566, USA}
\affil[3]{Faculty of Forestry, Hasanuddin University, Makassar, South Sulawesi, 90245, Indonesia}
\affil[4]{Department of Forestry, Faculty of Agriculture, Universitas Riau. Jl. HR Subrantas Km 12.5, Kampus Bina Widya, Riau, Indonesia}
\affil[5]{University of San Francisco, 2130 Fulton Street, San Francisco, CA 94117, USA}
\affil[*]{Address correspondence to: mwafiq@sig-gis.com, peter.cutter@recoftc.org, apoortinga@sig-gis.com}
\affil[$\dag$]{These authors contributed equally to this work.}
\date{}

\begin{document}
\maketitle

\begin{abstract}
Oil palm cultivation remains one of the leading causes of deforestation in Indonesia. To better track and address this challenge, detailed and reliable mapping is needed to support sustainability efforts and emerging regulatory frameworks. We present an open-access geospatial dataset of oil palm plantations and related land cover types in Indonesia, produced through expert labeling of high-resolution satellite imagery from 2020–2024. The dataset provides polygon-based, wall-to-wall annotations across a range of agro-ecological zones and includes a hierarchical typology that distinguishes oil palm planting stages as well as similar perennial crops. Quality was ensured through multi-interpreter consensus and field validation. The dataset was created using wall-to-wall digitization over large grids, making it suitable for training and benchmarking both conventional convolutional neural networks and newer geospatial foundation models. Released under a CC-BY license, it fills a key gap in training data for remote sensing and aims to improve the accuracy of land cover types mapping. By supporting transparent monitoring of oil palm expansion, the resource contributes to global deforestation reduction goals and follows FAIR data principles.
\end{abstract}

\section{Background \& Summary}
Oil palm is a highly valuable commodity globally and forms the backbone of numerous food and industrial supply chains \citep{zaki2025impact}. However, in Indonesia, the world's largest producer, oil palm cultivation has been a dominant driver of deforestation \citep{gunarso2013oil,austin2019causes,abood2015relative}. This transformation has far-reaching consequences: the use of fire for land clearing not only contributes to severe deforestation but also releases vast quantities of smoke and particulate matter, regularly deteriorating air quality across Southeast Asia and posing health risks to millions \citep{tornorsam2024managing}. While profitable, ecologically, oil palm expansion leads to habitat fragmentation, declines in biodiversity, and disruption of forest-dependent species \citep{feintrenie2010farmers}. In peat-rich areas, large-scale drainage for plantation establishment accelerates peat oxidation, contributing to widespread land subsidence—undermining long-term land productivity and increasing flood risk \citep{hooijer2012subsidence,ikkala2021peatland}. 

While numerous oil palm maps already exist \citep{descals2024global,clinton2024community,xu2020annual} most are based on point-level reference data and rely on pixel-based classifiers such as Random Forest (RF) model (e.g., \cite{uddin2021regional, teck2023land} and \cite{Weerakitikul2025}), which have limitations to meet the precision demands of existing and emerging compliance frameworks.  These frameworks require consistent, spatially explicit, and verifiable maps that can be audited and traced back to robust evidence. The advent of convolutional neural networks (CNNs) model \citep{poortinga2021mapping} and foundation models (FMs) has opened new possibilities for high-resolution land cover types mapping, offering significant improvements in spatial coherence and land cover type discrimination \citep{strong2025digital}. However, the uptake of these models is constrained by the scarcity of open, high-quality spatial datasets that are curated for benchmarking model performance or for training models to produce regulatory-grade maps \citep{xiao2025foundation}. What is needed is a new generation of benchmark datasets that provide plot-level training patches with full polygonal annotation, paired with rigorous typologies and Quality Assurance (QA)/Quality Control (QC) Protocols, to support the development and validation of next-generation Geospatial Artificial Intelligence (GeoAI) models.

Recent advances in GeoAI, particularly the use of CNNs and FMs trained on large-scale Earth Observation data, can significantly enhance the accuracy and scalability of land cover classification \citep{jakubik2023foundation}. These models excel at capturing spatial patterns and context, allowing them to distinguish complex land cover types such as mature versus immature oil palm, or oil palm versus visually similar tree crops. FMs also offer promising transferability across geographies and sensors. However, their performance is tightly coupled to the quality and structure of training data. Critically, there remains a shortage of spatially explicit, high quality, human-annotated datasets that are openly available. As a result, despite algorithmic advances, the lack of well-curated ground truth data continues to be a limiting factor in operationalizing these models for oil palm monitoring.

In particular, comprehensive “wall-to-wall” image labeling is essential for training modern machine learning (ML) models in RS. Deep learning models, including CNNs, often require fully segmented images (every pixel labeled) to learn context and spatial relationships, something that point samples or sparse labels cannot provide \citep{zhao2016learning,Boutayeb2024,Elharrouss2025}. Without dense annotations covering entire scenes, training signals are weaker and models may misinterpret surrounding landscape context. Wall-to-wall labeling ensures that models see both the target land cover type and all confounding land cover types in an area, improving their ability to differentiate oil palm from other land uses. In a recent study,  \citet{chaya2024cambodia} demonstrated that CNN's can distinguish between tree crops like cashew, rambutan and mango, which look very similar on moderate resolution satellite imagery, using wall-to-wall image labels. Fully annotated imagery provides richer information on texture and context, enabling models to learn subtle differences between land cover types. In contrast, traditional sample-based reference data (e.g., field plots or points) provide spatially sparse observations that lack continuous spatial context—limiting their utility in model training and validation \citep{radoux2020validation, chan2025sparse}.
 
Data needs are also increasingly driven by emerging regulatory and reporting frameworks. Under the United Nations Framework Convention on Climate Change, countries are required to submit transparent and verifiable Forest Reference Emission Levels and report progress toward their Nationally Determined Contributions. Similarly, global efforts such as the FAO’s Forest Resource Assessment and results-based finance mechanisms under REDD+ depend on consistent, spatially explicit land cover types and land uses data to ensure environmental integrity and track mitigation outcomes. The recent enforcement of the EU Deforestation Regulation introduces further demand for traceable, high-quality, and auditable maps to verify commodity sourcing and demonstrate compliance with zero-deforestation commitments \citep{berger2025earth}.

At the same time, commercial data providers are increasingly entering this space, offering proprietary classification layers and monitoring platforms. While these tools can be powerful, they often lack transparency, are not officially recognized by governments, and do not contribute to national data sovereignty. Moreover, their outputs are frequently non-replicable, limiting their utility in public regulatory contexts and raising ethical concerns around accountability, accessibility, and inclusivity. For jurisdictions with high levels of smallholder farming, it is essential that data, models, and outputs remain open, transparent, and reproducible—to ensure that these communities are not excluded or disadvantaged by opaque systems and inaccessible technology \citep{rist2010livelihood}. A publicly available, government-recognized benchmark dataset is thus critical for supporting equitable and sustainable land management at national and international scales.

In this study, we present a new open benchmark dataset for oil palm mapping in Indonesia, developed under the Lacuna Fund’s Climate and Forests 2023 initiative, with support from the Deutsche Gesellschaft für Internationale Zusammenarbeit (GIZ) on behalf of the German Federal Ministry for Economic Cooperation and Development. This contribution aligns directly with Lacuna Fund’s mission to reduce bias and fill data gaps in ML applications that benefit underserved populations in low- and middle-income countries. It supports GIZ’s FAIR Forward programme objective of fostering inclusive AI ecosystems and enabling equitable access to the benefits of AI technologies across partner countries.

Key contributions of this work include:

\begin{itemize}
    \item A high-quality, publicly accessible dataset, enabling robust and reproducible oil palm classification across diverse agro-ecological zones in Indonesia.
    \item Expert-labeled reference data, validated using high-resolution imagery and grounded in domain-specific knowledge, to support training and evaluation of ML models.
    \item Design for benchmarking, providing standard splits and protocols for comparative model evaluation across different methods and sensor combinations.
    \item Open licensing and documentation, ensuring broad accessibility, reproducibility, and alignment with Findability, Accessibility, Interoperability, and Reusability (FAIR) data principles.
\end{itemize}

\section{Methods}

\subsection{Study Area}

\begin{figure}[H]  
  \centering
  \includegraphics[width=1.0\linewidth]{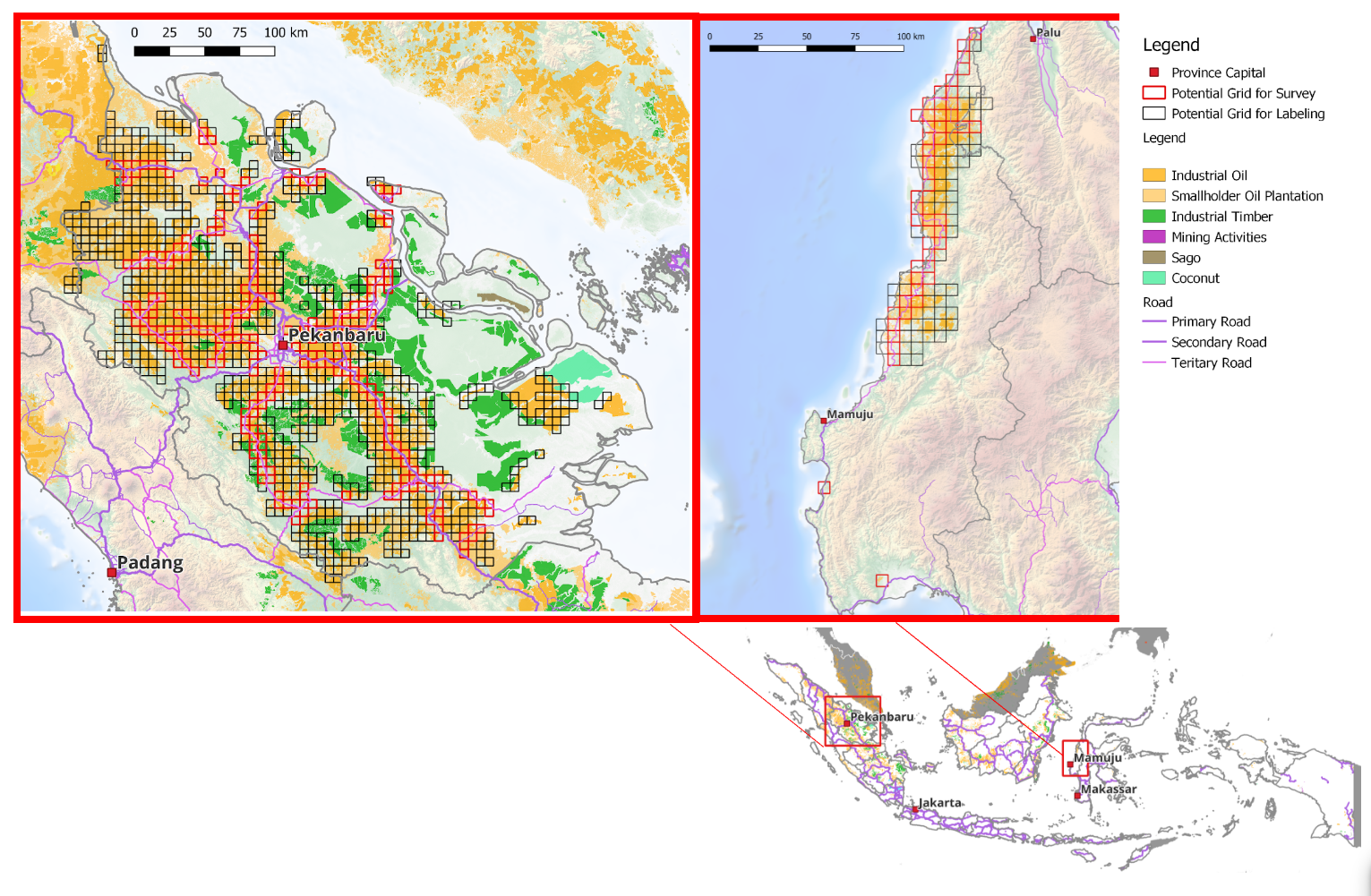}
  \caption{The two geographic regions of interest located in Riau (Sumatra) and West Sulawesi. The boxes indicate the potential sample cells.}
  \label{fig:studyarea}
\end{figure}

This reference dataset was collected in two provinces of Indonesia: Riau in Sumatra and West Sulawesi on the island of Sulawesi, as shown in Figure \ref{fig:studyarea}. These provinces represent ecologically and socioeconomically distinct oil palm production landscapes. Riau is a leading producer of crude palm oil in Indonesia, with over 2.5 million hectares of oil palm plantations as of 2020. The region has experienced extensive land use change over the past two decades, transitioning large tracts of forest and mixed agro-forestry into industrial and smallholder oil palm \citep{euler2016oil}. Its lowland topography and extensive plantation infrastructure make it an important area for reference data collection and model calibration. In West Sulawesi, the area of oil palm plantations increased from 61,159.95 hectares in 2015 to 152,763 hectares in 2019, indicating rapid oil palm expansion \citep{Anas2023}. The province features more heterogeneous landscapes with interspersed smallholder plots, forest fragments, and mixed perennial cropping systems. These attributes make it particularly valuable for assessing classification challenges in fragmented and transitioning regions \citep{Bainta2020,Aulia2020}.

\subsection{Land Cover Typology}

We constructed a hierarchical land cover typology to ensure consistent, interpretable annotation across all digitized areas. The schema balances visual separability in satellite imagery with thematic relevance to oil palm mapping and is summarized in Table \ref{tab:typology}.

At the highest level, we distinguish five broad categories: cropland, forest, shrubland, built-up, and other. These are refined at Level 2 to capture key land use and confusion sources. Within cropland, for example, we include perennial systems such as oil palm, rubber, coconut, rice field, and cacao, as well as seasonal or mixed other crops. Forest comprises intact or regenerating natural forest and, when present, mangrove/wetland forest elements. Shrubland captures sparse or early successional woody vegetation. Built-up covers settlements and infrastructure, and other includes non-vegetated or ambiguous land cover types such as bare ground and water.

Because oil palm is the main focus of this study, it was further subdivided (Level 3) by establishment and maturity stage to reflect substantial changes in appearance:

\begin{itemize}
    \item Palm (Initial planting): Areas under active establishment, often recently cleared or with very young palms and lacking canopy closure.
    \item Palm (Mature): Established plantations with closed, uniform canopy, regular texture, and darker green tones; inferred year of establishment was recorded when resolvable from time-series imagery.
\end{itemize}

Other perennial crops and natural/non-vegetated land cover types provide essential context and reduce misclassifications by supplying negative and confounding examples.

When interpreters were unable to assign a definitive land cover type, annotators used an "unknown" label within the appropriate branch (e.g., unknown perennial) and flagged those polygons for review, preserving ambiguity while maintaining traceability.

\begin{table}[H]
\centering
\caption{Hierarchical land cover typology used for annotation. Level 1 shows broad categories, Level 2 shows thematic subdivisions, and Level 3 provides detailed definitions or distinguishing characteristics.}
\label{tab:typology}
\hspace*{-2.5cm} 
\begin{tabular}{p{2.9cm} p{3.9cm} p{10.5cm}}
\toprule
\textbf{Level 1} & \textbf{Level 2} & \textbf{Level 3 / Descriptions} \\
\midrule
\multirow{9}{*}{Cropland} & \multirow{3}{*}{Oil Palm} & Palm (Initial planting): recently cleared or establishing; canopy not closed. \\
 &  & Palm (Mature): closed, uniform canopy with regular texture and darker green tone; inferred year of establishment recorded when resolvable. \\
 & Rubber & Dense canopy, irregular spacing, taller trees, gray-green tone. \\
 & Coconut & Widely spaced circular crowns. \\
 & Cacao& Scattered pattern, often under shade or mixed with other vegetation.\\
 & Rice Field& Seasonal rice paddies, somtimes with flooded appearance.\\
 & Other Agricultural Field& Ambiguous or low-certainty perennial crop; flagged for review.\\
\midrule
\multirow{2}{*}{Forest} & Natural Forest & Intact or regenerating forest with heterogeneous canopy and no plantation pattern. \\
 & Mangrove / Wetland Forest & Water-influenced structure with distinct morphology (tidal indicators) when present. \\
\midrule
\multirow{1}{*}{Shrubland} & Shrub / Grassland & Discontinuous low woody vegetation; early regrowth, degraded, or fallow. \\
\midrule
\multirow{1}{*}{Built-up} & Artificial surface & Rooftops, roads, infrastructure, and impervious surfaces. \\
\midrule
\multirow{4}{*}{Other} & Barren / Bareland& Non-vegetated exposed soil or cleared ground. \\
 & Water & Open water bodies with characteristic spectral signatures. \\
 & Unknown & Low-certainty non-vegetated/ambiguous cover; flagged for review. \\

\bottomrule
\end{tabular}
\end{table}

\begin{figure}[H]
    \centering
    \includegraphics[width=\linewidth]{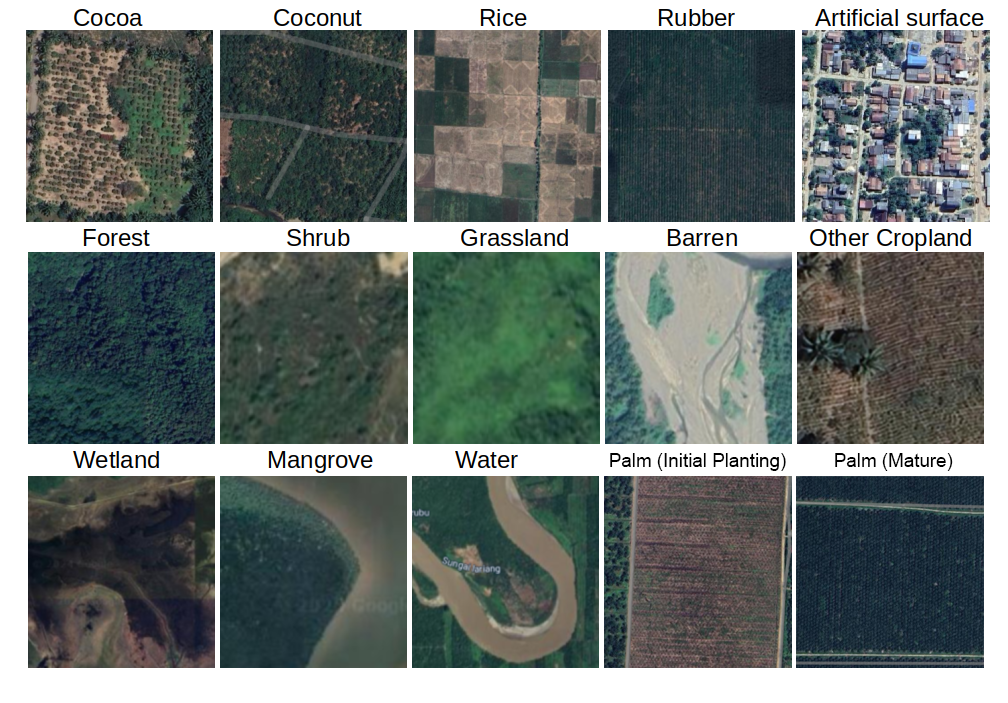}
    \caption{Land cover types of the study area.}
    \label{fig:classes}
\end{figure}

\subsection{Imagery and Temporal Reference}

Interpretation was conducted using a combination of high- and medium-resolution satellite imagery accessible via Collect Earth Online (CEO) and QGIS. The primary sources included:

\begin{itemize}
    \item \textbf{Planet NICFI Basemaps}: Monthly 4.7 m resolution imagery from 2020 to 2023, used as the principal source for detecting oil palm canopy structure, plantation maturity, and recent land use changes.
    \item \textbf{Sentinel-2}: 10 m resolution multispectral imagery (2017--2023), used for interpreting spectral patterns over larger time frames and corroborating visual features across seasons.
    \item \textbf{Bing and Google Satellite}: High-resolution true-color imagery (sub-meter where available), used primarily for contextual cues and manual annotation in areas with limited recent data coverage.
\end{itemize}

Image dates were carefully logged for each labeled feature, linking land cover type attribution to specific imagery dates. This ensured that all labels could be referenced to the visual evidence used and allowed traceability for model training and time-sensitive analysis.

Limitations included persistent cloud cover in some regions (especially West Sulawesi), seasonal variation in vegetation tone, and variable availability of high-quality imagery across years. Annotators were instructed to flag areas with poor image quality or temporal gaps for review.

\subsection{Data Collection and Digitization Approach}

\begin{figure}
    \centering
    \includegraphics[width=1.0\linewidth]{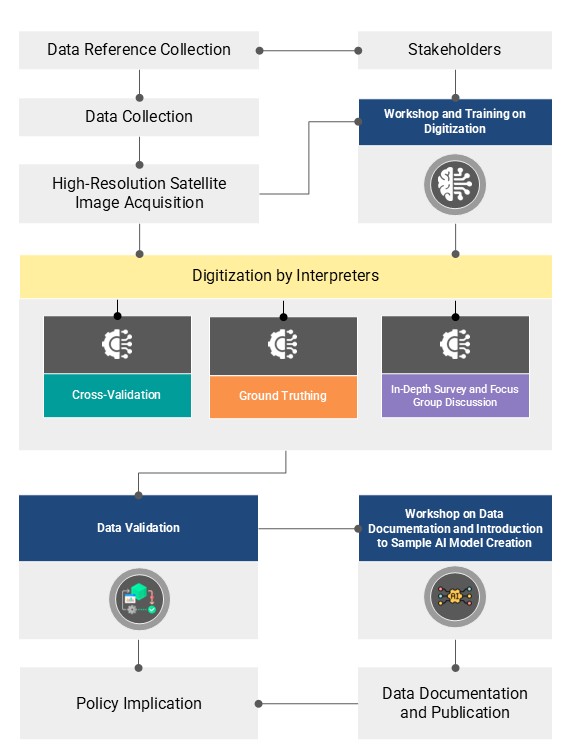}
    \caption{Reference data collection workflow of the open benchmark dataset for oil palm.}
    \label{fig:workflow}
\end{figure}

The dataset was produced through a structured digitization campaign. Figure \ref{fig:workflow} shows the data collection workflow. We began by overlaying a regular 6×6km grid over the study region and then applied an opportunistic selection strategy to choose grid cells with consistently high-quality, cloud-free imagery for manual annotation.

To define annotation targets, we applied a regular 6 × 6 km grid across the study area, consistent with the use of 6 × 6 km primary sampling units for regional spatial control as described by \citep{Lunetta2004}. Grid cells were then selected through a staged, opportunistic process. First, we filtered for cells overlapping areas known or expected to contain oil palm, based on existing maps, concession boundaries, and expert knowledge. Next, we retained only those cells with high-quality, cloud-free imagery from the past three years, using sources such as Planet NICFI mosaics, Sentinel-2 Level-2A surface reflectance (SR) with Cloud Score+ filtering, and Landsat 8/9 SR composites. Finally, the remaining candidates were reviewed by experts to ensure a diverse geographic and thematic sample, with priority given to challenging or underrepresented landscapes, including smallholder–forest mosaics, mixed cropping systems, and peatland regions. For each cell, trained interpreters produced wall-to-wall polygon annotations of all homogeneous land cover type patches.  

All interpreters were trained through a structured program run in partnership with Indonesian universities under the Ministry of Higher Education internship initiative. The curriculum included hands-on use of QGIS and CEO, application of the custom interpretation key, explicit decision rules, and best practices for polygon digitization \citep{Saah2019}. Regular calibration exercises, supervisor spot-checks, and feedback loops were used to align interpretations across contributors and reduce variability, ensuring the resulting annotations were thematically consistent and suitable for ML. 

\subsection{Quality Assurance Procedures}

We implemented a multi-stage QA/QC pipeline during and after annotation to ensure consistency and reliability. First, approximately 25\% of grid cells were double-annotated to enable cross-validation. Overlapping interpretations were compared using CEO’s built-in agreement metrics; discrepancies were resolved through supervisor review and group consensus, with decision rules and examples updated iteratively to align labeling. 

Interpreters flagged uncertainty via a confidence slider and free-text notes. Low-confidence or ambiguous polygons were either reviewed by senior annotators or temporarily labeled “Unknown” pending clarification; all such flags and confidence scores are retained in the metadata so downstream users can filter, weight, or exclude them. Senior team members also conducted routine spot-checks and thematic audits using QGIS and CEO, discussing difficult cases in weekly coordination meetings and updating the interpretation key to reflect refined definitions.

In addition to image-based review, we incorporated field validation in selected areas using Global Positioning System points, photographs, and participatory mapping data collected by local partners. These ground observations helped resolve ambiguities and were used to correct and document discrepancies in the mapped annotations. The final outputs were then uploaded to a centralized PostGIS database and passed through automated validation routines that checked for geometry errors, missing or inconsistent attribute codes, overlapping or duplicate features, and appropriate handling of very small polygons. This structured pipeline—from initial annotation through field verification and automated quality checks—ensured that all outputs were traceable, consistent, and suitable for downstream use as a high-fidelity polygon dataset.

\section{Validation}

In this study, we validated the land cover types wall-to-wall using a labeled dataset consisting of 52,225 polygons in fifteen land cover types, covering a total mapped area of 259,991.12 hectares. Oil palm dominates the dataset more than any other land cover type, with 12,598 polygons (139,488.49 ha) representing palm (mature) and an additional 3,719 polygons (15,155.33 ha) representing palm (initial planting), followed by other land cover types as shown in Table 2.

The original land cover types are divided into six main categories: (i) palm (mature and initial  planting), (ii) built up (artificial surface and bareland), (iii) cropland (other agricultural fields, coconut, rice field, rubber and cacao), (iv) natural forest (forest, mangrove and wetland), (v) shrubland (shrubland and grassland), and (vi) other (water, barren and unknown). This aggregation reduced land cover type-level confusion and enabled a more robust comparison of land cover type mapping.

To ensure the reliability of the reference dataset, validation was conducted using CEO through visual interpretation of very high-resolution satellite imagery. A stratified random sample of 4,800 plots (50 m × 50 m) was distributed across the study area, with one sample point collected per plot to capture representative land cover type conditions. The square plot design was chosen to ensure consistent interpretation of homogeneous land cover type patches. Furthermore, to assess interpreter consistency, 25\% of the plots were deliberately overlapped and re-assigned to multiple interpreters, enabling cross-checking and evaluation of agreement. This validation effort was supported by 30 trained student interpreters from Universitas Riau and Hasanuddin University.

\begin{table}[H]
\centering
\caption{Polygon counts and area by land cover type.}
\label{tab:polygon_counts}
\renewcommand{\arraystretch}{1.2}
\begin{tabular}{lcccc|c}
\hline
\textbf{Land cover type} & \textbf{Polygon Count} & \textbf{Area (Ha)} \\
\hline
Palm (Mature)            & 12,598  & 139,488.49 \\
Forest                   & 1,383  & 54,384.56 \\
Shrubland                & 6,795  & 25,705.09 \\
Water body               & 2,046  & 16,501.32 \\
Palm (Initial planting)  & 3,719  & 15,155.33 \\
Bareland                 & 5,656  & 8,969.44 \\
Other agricultural field & 2,186  & 5,734.26 \\
Built Up Area            & 12,521 & 4,494.77 \\
Coconut                  & 1,138  & 4,227.55 \\
Rice Field               & 2,903  & 3,223.49 \\
Grassland                & 488    & 2,864.76 \\
Rubber                   & 232    & 1,512.11 \\
Wetland                  & 304    & 1,114.50 \\
Cacao                    & 206    & 438.45 \\
Mangrove                 & 50     & 196.75 \\
\hline
\textbf{Total}           & \textbf{52,225} & \textbf{259,991.12} \\
\hline
\end{tabular}
\end{table}

The study applied an accuracy assessment to validate the land cover type mapping, using PA, UA, OA, and Kappa for six main categories across 72 grids. This validation aligns with prior research by \citet{poortinga2019mapping}, \citet{Cheng2019}, \citet{Xu2020} and \citep{danylo2021oilpalm} in mapping plantations (e.g., land cover type classification and oil palm mapping), which underscores the necessity of accuracy assessment for reliable mapping and change detection analysis, as formulated by \citep{Stehman2009, Olofsson2014, olofsson2020mitigating}.

The overall classification results indicated a strong performance with an OA of 0.83 and a Kappa of 0.76, suggesting substantial agreement beyond chance. The oil palm land cover type, which was central to the mapping objective, achieved the highest classification quality, with a PA of 0.96 and UA of 0.88. This indicated both high recall and precision, affirming the reliability of the annotated dataset for detecting palm oil plantations, especially mature stands. Other well-performing land cover types included built-up (UA = 0.86, PA = 0.83) and natural forest (UA = 0.84, PA = 0.87), both of which exhibited distinct spatial and spectral signatures, likely contributing to their separability in classification.

In contrast, the cropland and shrubland land cover types demonstrated lower classification accuracy, with Cropland showing particularly poor performance (UA = 0.23, PA = 0.58). This was likely due to spectral confusion with adjacent land uses, such as young oil palm or mixed smallholder agroforestry, which may not be easily distinguishable in medium-resolution imagery. Shrubland also had modest accuracy (UA = 0.74, PA = 0.51), suggesting difficulty in delineating it from degraded forest or young regrowth. These findings highlighted priority areas for improving annotation precision or refining land cover type definitions, particularly in transitional or mixed-use landscapes. Continued emphasis on inter-annotator consistency and enhanced field validation in these complex land cover type could further improve the overall robustness of the classification system.

\begin{table}[H]
\centering
\caption{Accuracy assessment of the land cover types. 
PA stands for Producer's Accuracy, UA stands for User's Accuracy, 
OA stands for Overall Accuracy, and Kappa stands for Kappa Coefficient.}
\label{tab:accuracy_lulc}
\renewcommand{\arraystretch}{1.2} 
\resizebox{\textwidth}{!}{%
\begin{tabular}{lcccccc|c}
\hline
\multicolumn{8}{c}{OA = 0.83 \quad Kappa = 0.76} \\
\hline
\textbf{Land cover type} & \textbf{Oil Palm} & \textbf{Built-up} & \textbf{Cropland} & \textbf{Natural Forest} & \textbf{Shrubland} & \textbf{Other} & \textbf{UA} \\
\hline
Oil Palm        & 2445 & 32  & 19  & 18  & 147 & 118 & 0.88 \\
Built-up        & 5    & 206 & 1   & 0   & 6   & 22  & 0.86 \\
Cropland        & 59   & 2   & 49  & 18  & 57  & 32  & 0.23 \\
Natural Forest  & 5    & 1   & 4   & 802 & 131 & 16  & 0.84 \\
Shrubland       & 10   & 4   & 10  & 76  & 388 & 36  & 0.74 \\
Other           & 18   & 3   & 2   & 11  & 26  & 531 & 0.90 \\
\hline
\textbf{PA}     & 0.96 & 0.83 & 0.58 & 0.87 & 0.51 & 0.70 &  \\
\hline
\end{tabular}
}
\end{table}

\section{Data Records}
The oil palm reference dataset, titled Open Land Use Reference Dataset for Palm Oil Landscapes in Indonesia \citep{Wafiq2025}, is available in Zenodo under a CC-BY 4.0 license, ensuring it is free to use with attribution. The dataset consists of geospatial files containing the annotated polygons and their attributes, along with accompanying metadata and documentation. Table 1 (in the Methods section) provides the classification schema used, and the data files include this information in structured form. Repository and files: All data have been deposited in Zenodo and can be accessed via DOI: https://zenodo.org/records/15618532. The core data are provided as a vector GIS file (Esri Shapefile and GeoJSON formats are available for convenience) containing all labeled polygons. We also provide the data in a Geopackage format, which includes all attributes in one file with internal topology. To facilitate use in Google Earth Engine (GEE) or similar platforms, a simplified CSV with geometry references or an Earth Engine Asset are also provided. 

A detailed README file and the interpretation key (as a PDF manual) are included to guide users in understanding the land cover type definitions and methodology. Spatial characteristics: The polygon dataset covers selected 6x6 km grid cells in Riau and West Sulawesi provinces. In total, the dataset contains on the order of a few thousand polygons, spanning an area of several hundred square kilometers of annotated land. Each polygon delineates a homogeneous land cover type patch as of the 2020–2023 period. The coordinate reference system of the data is WGS 84 (EPSG:4326) for compatibility with most global mapping tools, and coordinates are in latitude/longitude. The average polygon size varies by land cover type (e.g., individual smallholder plots might be 1–5 hectares, while a single mature plantation polygon could be tens of hectares if it’s a contiguous estate block; meanwhile small features like isolated houses or ponds are mapped as small polygons if distinguishable). Adjacent polygons share boundaries with no gaps or overlaps beyond minimum mapping unit, creating a seamless coverage within each grid cell.

\section{Usage Notes}

This open benchmark dataset supports multiple applications in RS and land-use science:

1. \textbf{Model training:} Wall-to-wall, high-resolution polygon labels enable supervised learning for dense land cover type mapping, including segmentation models and FMs. The hierarchical typology supports multi-scale or multi-task formulations, and the data can be used to fine-tune pre-trained GeoAI models for improved oil palm discrimination \citep{wiratama2025comparative}.

2. \textbf{Benchmarking and transfer evaluation:} Standard splits facilitate fair comparison of methods (e.g., RF, U-Net, Vision Transformer) and assessment of spatial transfer (e.g., models trained in Riau versus West Sulawesi) \citep{papoutsis2023benchmarking}.

3. \textbf{Product validation:} The dataset serves as ground truth for validating existing land cover products, concession maps, or global datasets (e.g., forest loss maps), including evaluation of maturity-stage classification. Users should account for the opportunistic sampling design when interpreting area-based accuracy metrics \citep{Olofsson2014}.

4. \textbf{Land use change analysis:} Annotated conversion years for many oil palm plots allow temporal trend analysis, estimation of expansion timing, and distinction between forest conversion and other land use transitions—informing carbon, biodiversity, and policy monitoring \citep{li2018oilpalm}.

5. \textbf{Comparative landscape studies:} Inclusion of other crop types and natural covers enables analysis of plantation configurations, fragmentation, and socio-environmental differences between smallholder and industrial systems, especially when combined with ancillary layers (e.g., concessions, peat maps) \citep{descals2019oilpalm}.

6. \textbf{Fairness and inclusivity in AI:} Data from diverse, under-represented Indonesian contexts can be used to reduce geographic bias in global models and bootstrap similar efforts in other regions via transfer learning \citep{ferrara2024fairness}.

7. \textbf{Capacity building and education:} The openly licensed, well-documented data are suitable for teaching RS and ML workflows, offering local stakeholders hands-on tools for analysis and advocacy \citep{huang2025application}.

8. \textbf{Monitoring/near real-time (NRT) applications:} Models trained on this static reference can be deployed with updated imagery streams (e.g., Sentinel-2) to track expansion or flag potential illegal conversion, with maturity metadata supporting yield and production forecasts \citep{oconnor2020earth}.

\textit{Practical considerations:} The vector data are moderate in size (tens of MB); users training pixel-based models should rasterize polygons to match sensor resolution (recommended 5–10 m). land cover type imbalance (e.g., oil palm over-represented by design) may necessitate weighting or balancing strategies. Because sampling is opportunistic, the dataset is best used for pattern learning, benchmarking, and relative comparisons rather than direct area estimation without corrective sampling design.

\section{Code Availability}

To support transparent use and reproducibility, we provide both the dataset and a curated set of scripts through our public repository on Zenodo. The repository includes all polygon annotations, metadata, and supporting materials, along with selected utility scripts for data preprocessing and analysis. 
All supporting materials—such as SQL queries for PostGIS processing, 
polygon QA/QC scripts, labeling guides, and sample analysis notebooks—
are available in the dataset repository \citep{Wafiq2025}. These include:

\begin{itemize}
  \item Scripts for geometry and attribute validation:
  \item Templates and resources used in interpreter training (e.g., QGIS styles, decision rules):
  \item Accuracy assessment notebooks, including confusion matrix computation and visualization:
  \item Export routines for preparing raster labels or Earth Engine–compatible datasets:
  {\footnotesize
    \begin{itemize}
      \item \textbf{Dataset Visualization}: 
        {\scriptsize \href{https://code.earthengine.google.com/58bdfbf5ebc2549df2b04f3d1b5280f4}{https://code.earthengine.google.com/58bdfbf5ebc2549df2b04f3d1b5280f4}}
        
      \item \textbf{Landsat $8/9$ (30 m)}: 
        {\scriptsize \href{https://code.earthengine.google.com/af03f5bc263a897c5b02f73609478b81}{https://code.earthengine.google.com/af03f5bc263a897c5b02f73609478b81}}
        
      \item \textbf{Sentinel-2 ($10/20$ m)}: 
        {\scriptsize \href{https://code.earthengine.google.com/bfdb2891feb75b14f7cd6e5593e76cc2}{https://code.earthengine.google.com/bfdb2891feb75b14f7cd6e5593e76cc2}}
    \end{itemize}
  } 
\end{itemize}

All materials are shared under a CC-BY 4.0 license, in line with the dataset.

\bibliographystyle{plainnat}
\section*{Acknowledgments}
This work was supported by the Lacuna Fund’s Climate and Forests 2023 initiative, with financial and technical support from GIZ on behalf of the German Federal Ministry for Economic Cooperation and Development (BMZ). We thank RECOFTC, the Spatial Informatics Group (SIG), and the Spatial Informatics Group–Natural Assets Laboratory (SIG-NAL) for their collaboration and contributions to dataset development. This effort aligns with the Lacuna Fund’s mission to reduce data gaps in ML for underserved regions and supports GIZ’s FAIR Forward programme in fostering inclusive and equitable AI ecosystems.

\bibliography{references}




\end{document}